%% file: egpaper_for_review.tex
\documentclass[10pt,twocolumn,twoside, letterpaper]{article}

\usepackage{times}
\usepackage{epsfig}
\usepackage{graphicx}
\usepackage{amsmath}
\usepackage{amssymb}
\usepackage{booktabs} %
\usepackage{epstopdf}
\usepackage{verbatim}
\usepackage{paralist}
\usepackage{multirow}
\usepackage{cleveref}
\usepackage{subfig}
\usepackage{balance}
\usepackage{fancyhdr}
\date{}
\newcommand{\norm}[1]{\left\lVert#1\right\rVert}

\def\wacvPaperID{****} %

\makeatother
\begin{document}

\title{CCCNet: An Attention Based Deep Learning Framework for Categorized Crowd Counting}

\author{Sarkar Snigdha Sarathi Das, Syed Md. Mukit Rashid, and Mohammed Eunus Ali \\Department of Computer Science and Engineering\\
Bangladesh University of Engineering and Technology\\
{\tt\small sarathismg, mukitrashid270596, mohammed.eunus.ali@gmail.com}
\and
~
}

\maketitle

\begin{abstract}
  \input{transfer/abstract.tex}
\end{abstract}

\section{Introduction}
\input{transfer/Introduction.tex}

\section{Related Works}
\input{transfer/related.tex}

\section{Our Approach}

\input{transfer/methodology.tex}

\section{Model Training}
\input{transfer/Training.tex}

\section{Experiments}
\input{transfer/Experiments.tex}

\section{Conclusion}
\input{transfer/Conclusions.tex}

{\small
\bibliographystyle{ieee}
\bibliography{egbib}
}

\end{document}

%% file: transfer/abstract.tex
Crowd counting problem that counts the number of people in an image has been extensively studied in recent years. In this paper, we introduce a new variant of crowd counting problem, namely \textbf{categorized crowd counting}, that counts the number of people sitting and standing in a given image. Categorized crowd counting has many real-world applications such as crowd monitoring, customer service, and resource management. The major challenges in categorized crowd counting come from high occlusion, perspective distortion and the seemingly identical upper body posture of sitting and standing persons. Existing density map based approaches perform well to approximate a large crowd, but lose important local information necessary for categorization. On the other hand, traditional detection-based approaches perform poorly in occluded environments, especially when the crowd size gets bigger. Hence, to solve the categorized crowd counting problem, we develop a novel attention-based deep learning framework that addresses the above limitations. In particular, our approach works in three phases: i) We first generate basic detection based sitting and standing density maps to capture the local information; ii) Then, we generate a crowd counting based density map as global counting feature; iii) Finally, we have a cross-branch segregating refinement phase that splits the crowd density map into final sitting and standing density maps using attention mechanism. Extensive experiments show the efficacy of our approach in solving the categorized crowd counting problem.

%% file: transfer/Introduction.tex
The crowd counting problem that counts the number of people in a given image, has gained considerable attention in recent years due to its intense demand in video surveillance, public safety, and urban planning. Counting crowd by automatic scene analysis is a challenging task due to occlusion, complex background, non-uniform distributions of scale and perspective variations. A plethora of techniques have been proposed in recent years
\begin{figure}[ht]
\begin{center}
\includegraphics[width=0.48\textwidth,height=0.12\textheight]{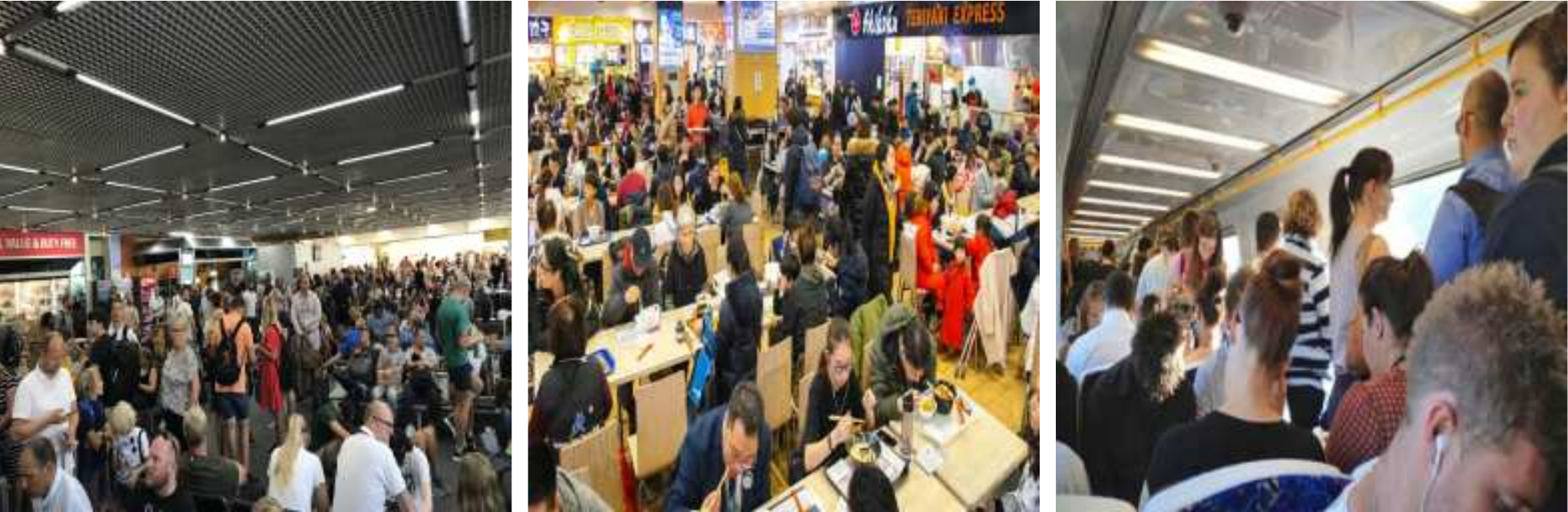}
\caption{Example images from our dataset}
\label{fig_ds}
\end{center}
\end{figure}

\noindent (e.g.,~\cite{ref_DivideandGrow,ref_crowdnet,ref_decidenet}) to address these challenges and to increase the accuracy of crowd count in different real-world environments.

In this paper, we introduce a new variant of crowd counting, namely \textit{categorized crowd counting}, that counts the number of persons sitting and standing separately in a given image. There are many practical applications of categorized crowd counting. For example, a bank manager may want to know the number of customers who are waiting, standing inside the service area of the bank so that s/he can increase the on-demand resource for better service to the customers; a bus/tram operator may want to know the number of standing passengers and sitting passengers in the bus/tram, which will help them to decide on the frequency and size of transports needed in different times of the day; a service provider may want to know the number of standing and sitting customers in a room to decide on the facility that they should provide. In general, the categorized crowd counting will add a new dimension in providing quality services especially in restaurants, banks, airport waiting areas, subway, and public transport where delivering quality customer service is crucial. To the best of our knowledge, we are the first to attempt the problem of categorized crowd counting.

Existing approaches for general crowd counting can be largely divided into two groups: (i) the most recent density-based approaches (e.g.,~\cite{ref_DivideandGrow,ref_crowdnet,ref_decidenet, valloli2019w, sam2019almost, hossain2019crowd}) that generate density of the crowd to approximate a large crowd in outdoor environment, and the detection based approaches that detect visible human body parts~\cite{ref_SummaryOfDetection,ref_Detection2} to count the number of persons in a given (mostly indoor) image. Though the density-based counting is quite promising when counting people in a high-density crowd, it has the following limitations: (i) For images with a low-density environment, the density-based approaches usually overestimate the crowd count; (ii) Images where global crowd features vary significantly due to the presence of obstacles between people (e.g., in lightly crowded indoor images with furniture such as tables and chairs), the performance of these global density-based counting methods drops significantly; and (iii) there is no way to differentiate between standing and sitting crowd as the local information corresponding to the persons present in the image are lost during the density map creation. On the other hand, detection-based approaches fail to address the categorized crowd counting due to the following limitations: (i) As the crowd density in images increases, the detection accuracy and the reliability of extracted local information start to decrease. The detection accuracy is poor in highly occluded images; and (ii) Even if it was possible to count the number of people by counting heads, it would be quite impossible to detect the state of the body due to occlusion.

To solve the categorized crowd counting problem, we propose a deep learning based approach that fuses both global (density of different parts of the image) and the local (state of the body) information present in the given image. The key idea of our approach comes from the following real-world observations: (i) Relative position of a person's body parts, their visibility and depth information of a human subject are important cues when differentiating a standing person from a sitting person, and (ii) the detection and labeling of some of the persons present would assist in categorization of other persons present in an image.

Our proposed solution framework consists of three major phases. First, we use a pose estimator to detect persons and extract pose features. From those features, we use a neural network based classifier to get a baseline classification of the detected persons with a weighted linear regressor used to further ameliorate the baseline classification. The output of this detection based baseline categorization is used to generate two basic density maps for sitting and standing persons respectively. This step utilizes the relative position of the persons' body parts and also the depth information for the classification of standing and sitting persons. In the next step, we generate a density map of the full crowd present in the image. We first generate a regression map for crowd counting via a CNN, and using this map and the density maps generated in the first phase, we generate a density map of the total crowd which is adaptive to varying crowd densities. Finally, we adopt a cross-branch segregating refinement phase that uses the detection based standing and sitting crowd map of the first phase and the crowd count map of the second phase to determine respective attention weights and produce the final sitting and standing density maps. Summation of these density maps gives us the final counts for each category.

To validate the effectiveness of our approach, we build a new dataset as existing benchmark datasets do not contain a sufficient number of sitting people and do not have ground truth annotations for each category. Our dataset contains 553 crowd images containing a total of 16521 people taken from a large variety of environments especially from places like restaurants, airport waiting areas, public transport, etc., where the categorized crowd counting problem is of much importance. Some example images of our dataset are shown in Figure \ref{fig_ds}. The dataset also contains images of varying densities ranging from 1 to 206 persons per image. Extensive experiments show the effectiveness of our approach in solving categorized crowd counting problem in a wide variety of real-world environment achieving an MAE of 4.15 and 4.80 and RMSE of 7.96 and 8.59 for sitting and standing crowd respectively, significantly outperforming the existing state-of-the-art traditional crowd counting schemes adapted to this multi-task problem.

In summary, our contributions are as follows:
\begin{compactitem}
\item We are the first to introduce the categorized crowd counting problem, which can assist in different applications such as crowd monitoring, customer service and, resource management.
\item  We propose a novel three-phase deep learning based approach for categorized crowd counting that exploits both local and global features with attention mechanism to count each category of persons independently.
\item  We conduct extensive experiments and show the effectiveness of our approach in various densities and cross-scene environments.
\end{compactitem} 

%% file: transfer/related.tex
Based on the working methodologies, existing works on crowd counting can be divided into three groups: counting by detection, counting by global regression, and counting by deep learning.\\

\noindent\emph{Counting by Detection: }The early approaches in crowd counting that count a small number of people in a given image were mainly based on different types of detection. Example includes non-maximum suppression based detection~\cite{ref_discriminative}, hough transform based detection \cite{ref_hough}, pedestrian detection \cite{ref_pedtraffic}, edgelet part detection \cite{ref_edgelet}, and combining local part based scheme with global shape template \cite{lin2010shape}. Relating object parts was another motivation of some of the approaches \cite{dong2007fast,wu2009detection}. 
\begin{figure*}[t]
\begin{center}
\includegraphics[width=0.95\textwidth,height=0.23\textheight]{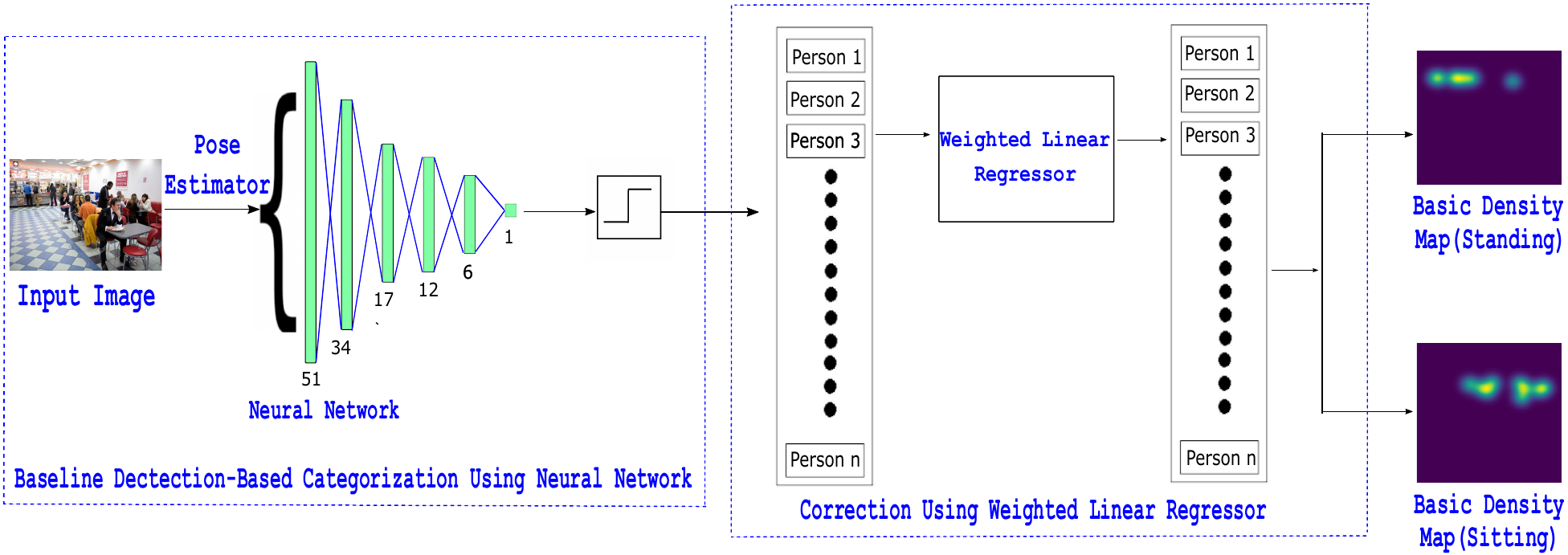}
\caption{\label{module1} A block diagram of detection based basic density map generation}
\end{center}
\vspace{-10mm}
\end{figure*}
\newline~\newline~
\emph{Counting by Global Regression: } In high crowd densities, using global level crowd features became the key to crowd counting. Kernel ridge regression \cite{an2007face},
multi output regression models \cite{chen2012feature}, blob size histograms to eliminate perspective problems \cite{kong2006viewpoint}, usage of different features from interesting points \cite{ref_Idrees} and random forest regression \cite{fiaschi2012learning} are some notable works. Some approaches also utilized both detection and regression methods to get an estimate of the crowd count based on segmentation \cite{chan2008privacy,ryan2009crowd}. These methods were prone to losing local information, which resulted in poor performance in the higher density crowd. To alleviate this problem, the work in \cite{ref_Lemptisky} pioneered a new way by creating a density map from a given image. Another new way of counting was introduced in \cite{arteta2014interactive} which was based on feedback response from user. This method was introduced when most of the methods at that time were completely unable to estimate count in a new scene.

\noindent\emph{Counting by Deep Learning: }As deep learning seemed to outperform traditional methods in different computer vision problems, the first time CNN was used for crowd counting in \cite{ref_cross} by optimizing density loss and crowd counting loss. In \cite{ref_single}, a multi column CNN was introduced to deal with density variation of different crowd images. An approach of directly mapping crowd images to the count was also taken \cite{wang2015deep}.  Walach et al. \cite{walach2016learning} worked with multiple CNNs where errors are corrected by subsequent networks. Multi scale input and fusion in last layer was used in \cite{onoro2016}. Other recent deep learning based approaches include switching mechanism between specialized columns \cite{ref_switch} and dynamic representation and appearance in crowd video understanding \cite{shao2016scnn,ref_DivideandGrow,zhang17sacnn}. Another contemporary approach used a CNN model emphasizing on head locations \cite{zhang2018attention}. Also CNN models targeted to generate low resolution density map first and estimating high resolution density map from the low resolution map has been addressed in~\cite{ranjan2018iterative}. These approaches primarily focus on highly dense crowd scenarios. Liu et al. \cite{ref_decidenet} found that, density map based approaches tend to overestimate the crowd count in lower density sparse crowd images, where the detection based methods generally work well. As categorized counting is more relevant in low to medium density images, traditional crowd counting techniques are likely to give overestimation in these environments. Moreover, density maps generated for crowd counting lose local information, which is essential for categorized crowd counting. To address this issue, they used a spatial attention model \cite{ref_decidenet}. Some recent techniques \cite{hossain2019crowd, kang2018crowd} have also used attention mechanism for crowd counting. \cite{valloli2019w} recently used U-Net like architecture which uses reinforcement branches to aid in counting, and achieves consistent state of the art results in majority of the datasets. While recent models do a reasonable job in estimating crowd count, emphasis only on global features makes these methods unsuitable for crowd categorization.

%% file: transfer/methodology.tex
A straightforward way to build a categorized counting framework is to create an end-to-end network for classifying sitting and standing people in an image. Due to the complex nature of the problem we found that an end-to-end network increases the difficulty of parameter tuning. Thus, we propose a three phase approach where in each phase it learns some set of specific information from the image and forwards the learned information to the next phase, thus eliminating the need to tune parameters of the whole network at once. In particular our framework works in three major phases, which are described as follows.
\begin{figure*}[t]
\begin{center}
\includegraphics[width=.87\textwidth,height=0.33\textheight]{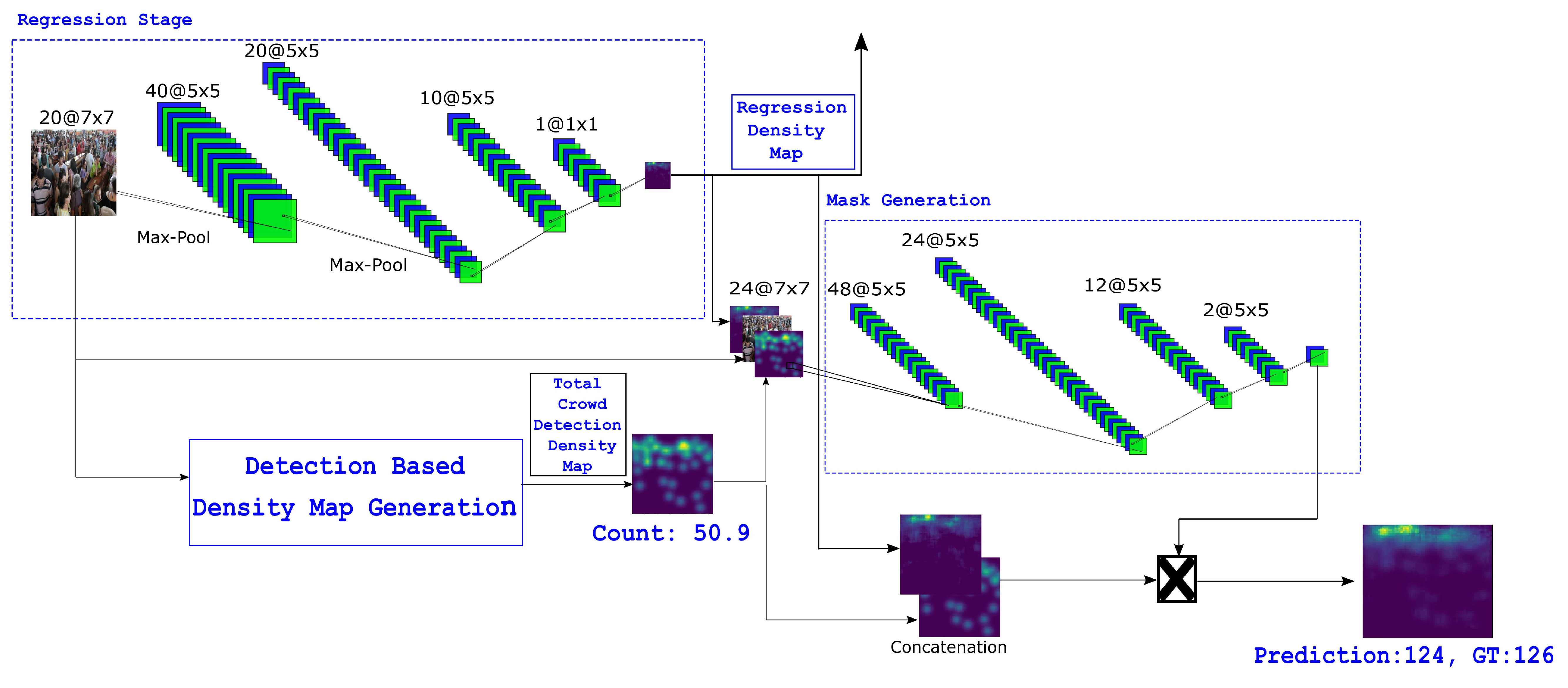}
\caption{A block diagram of crowd counting based density map generation. Dimensions and channels of the filter applied in each layer is shown on top of that layer. For example, on top of the input image, 20@7x7 denotes 20 filters with 7x7 dimension is applied on this layer.}
\label{module2}
\end{center}
\end{figure*}
\subsection{Detection Based Basic Density Map Generation} \label{sec:DNN}
In this phase, we first classify the body state of different persons based on the location of their detected body parts and generate two separate detection based basic density maps for sitting and standing people respectively.

Since the detection based techniques are proven to perform reasonably well in counting the number of people in a low density image, we first build a neural network model that can classify a detected person as standing or sitting. We first adopt a pre-trained multi-person pose estimator model~\cite{ref_alphapose}. It is a pose estimator network which incorporates a symmetric spatial transfer network (SSTN), a single person pose estimator (SPPE) and a parametric pose non-maximum suppression (Pose NMS) that detects individual persons and give locations of different body parts of all the detected persons in an image. For each person, 17 key body joints are detected (e.g. nose, shoulders,knee, ankles etc.). With 2 values for the locations and a value indicating the confidence with which the body part is located, we have a total of 51 outputs for each person. Next we develop a fully connected feed-forward neural network as shown in Figure \ref{module1} which takes the aforementioned 51 outputs for each person as input and obtains a baseline body state classification for that person. This network consists of 4 hidden layers with 34,17,12 and 6 nodes respectively.  The Adam optimizer is used to optimize the network parameters and binary cross-entropy is selected as the loss function. A baseline classification of all the detected persons in the image is obtained. Afterwards, a weighted linear regressor is used to further refine the baseline classification, which estimates a decision plane that separates sitting people from standing people. This linear regressor uses the location of the nose point as an estimation of the location of the person, and is weighted on the confidence of the identified body keypoints, which essentially indicates the visibility of the person.

As the final step of this phase, we produce separate crowd density maps for each category (sitting and standing). To generate the density maps from labeled persons with their nose point used as the annotation, we convolve the annotations with geometry-adaptive Gaussian kernels using the same technique as in \cite{ref_single}. This gives us the basic detection based density maps of the different categorizations. They only contain categorization of the persons detected by the pose estimator. However, the detection accuracy falls severely in higher density environments. To improve the total count estimation and categorization of persons in occluded regions, we use these density maps in the second and third phase of our pipeline.
\begin{figure*}[t]
\begin{center}
\includegraphics[width=0.87\textwidth,height=0.33\textheight]{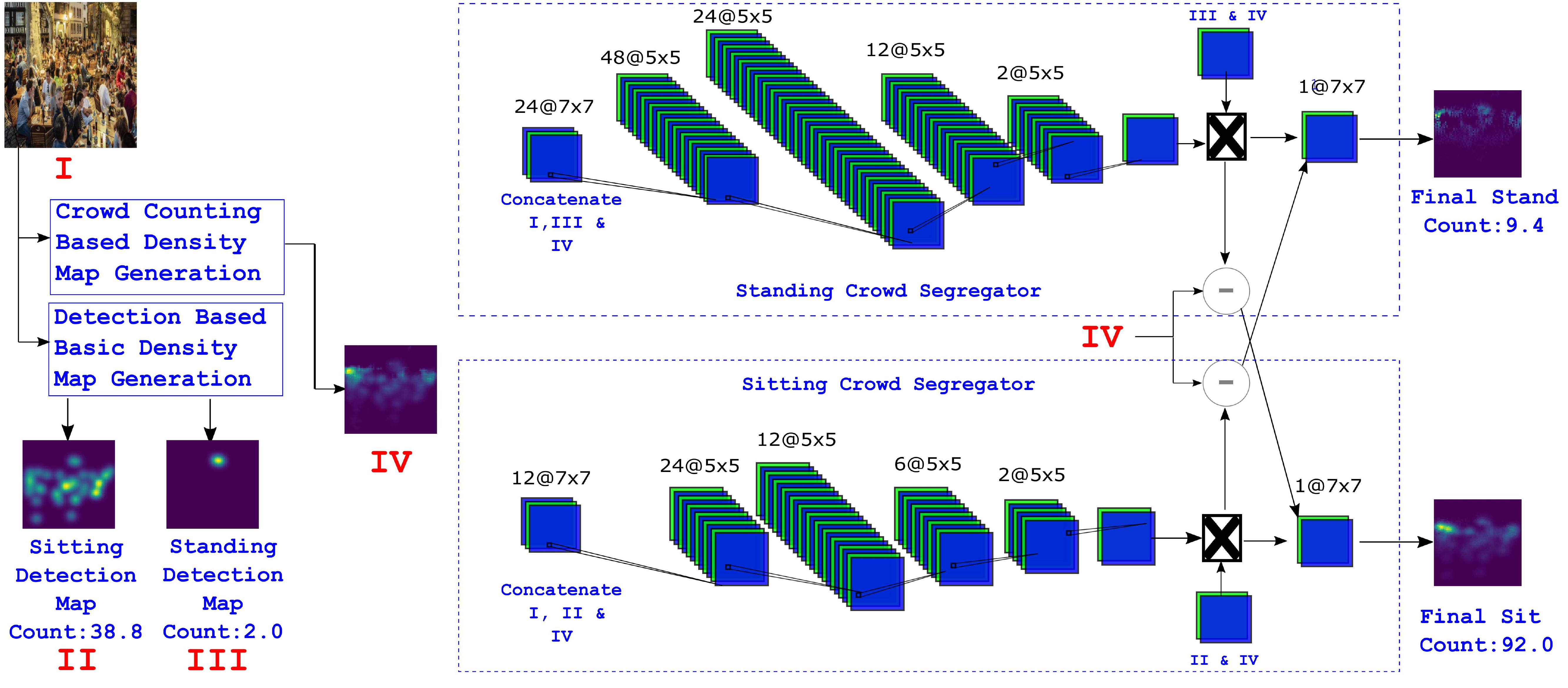}
\caption{The block diagram of cross-branch segregating refinement phase. Here II, III and IV denotes the sitting and standing detection maps from the first phase and the total crowd density map coming from the second phase respectively.}
\label{module3}
\end{center}
\end{figure*}
\subsection{Crowd Counting Based Density Map Generation}
In the second phase of our framework we generate a total crowd density map to capture the global information present in the image. Inspired by the success of \cite{ref_decidenet} to give an accurate count in varying densities, we exploit a similar idea to make a simpler crowd counting model. We first generate a crowd regression map, which is similar to the shortest field of view branch(starting with 5x5 filter) of the MCNN architecture proposed by \cite{ref_single}, where we infer a density map using several convolutional layers of the input image. The details of this part of the phase is given in Figure~\ref{module2}. Following the outputs of the first two layers, we use max-pooling to get features irrespective of various transformations.

Next we learn two masks that essentially determines the relative weights of initial crowd map and categorized maps in determining the final crowd map. The input image is down-sampled to one-fourth of the width and height.Then the crowd regression output, the down sampled image, and the crowd detection map are concatenated together. This stacked result is channelled through a 5 layer CNN architecture as shown in Figure \ref{module2}. Apart from the first layer, all the other layers use $5\times5$ kernel size, which works well in an environment prone to scale variance. This part results in two weight masks which are multiplied by the regression and detection based crowd density maps respectively and the products are sent through a single channel $1\times1$ filter to get the final crowd density map. Using two separate attention masks give much flexibility to the network to choose from noisy regression and detection maps.

\subsection{Cross-Branch Segregating Refinement}
While we obtain detection based basic density maps for sitting and standing in the first phase, the major drawback in that phase was the diminishing detection rate in images with higher person density due to extreme occlusion. To alleviate this problem, we combine the dense crowd information from the second phase with the initial detection results of the first phase to further refine the results of categorized crowd counting.\\

This phase encompasses 2 parallel network branches to obtain separate density maps for sitting and standing persons. Both the networks are fed the corresponding density map (Sitting / Standing), concatenated with the total crowd density map and input image. Then we add another 5 layers for each branch as shown in Figure \ref{module3}. This results in two different bi-channel segregation attention maps in each branch, which essentially gives the network a notion of how to divide the total crowd density map. So, we multiply these maps with corresponding detection density map and the crowd density map. But the branches have to agree in terms of this segregation so that same part of the image shouldn't be labelled as both standing and sitting. So, it has to be somehow incorporated in the loss.\\

This leads us to a concept of internal cross connection between these branches. The primarily refined sitting density map is subtracted from total count estimate, resulting in a subtracted standing map. This output is concatenated to the output of the other branch standing estimate. A 7x7 convolution in this step gives us the final standing density map. Similar strategy is followed to get the final sitting density map, as seen in Figure~\ref{module3}. This cross-connection makes both the branches accountable for the errors in the other branches too, restricting them from being biased towards a single side.

%% file: transfer/Training.tex
We train the three phases of our methodology independently which eliminates the difficulty of paramter tuning in an end-to-end model. There is another considerable advantage in choosing this approach. We can now use the Shanghaitech(part B) \cite{ref_single} dataset along with ours to train the second phase to make the counting strength of our model more robust, so that it can handle varying densities properly. 70\% of our dataset is used for training, 15\% for validation and the rest 15\% for testing. The same split is used throughout the whole procedure to avoid any kind of peeking.
\subsection{Detection Based Basic Density Map Generation}
The feed-forward neural network used after the pose estimation is quite simple in structure and easy to train. We run the pose estimator on the training and validation images and extract the body keypoints found of all the detected persons. Using the detected persons from these images as training and validation data respectively, we train the neural network model with Adam optimizer using an initial learning rate of $8\times {10}^{-3}$. The whole model is trained for 10000 epochs with a batch size of 512. The model which performed the best in the validation set was saved and used. In the weighted linear regressor used after this step, the weight of a sample is given by $ W = 2*{U} + {L}$ where $U$ and $L$ is the sum of all confidences of the upper 10 (two eyes, ears, shoulders, wrists and elbows) and lower 6 (two hips, knees and ankle) body part keypoints respectively.
\subsection{Crowd Counting Based Density Map Generation}
This part of the model is trained end-to-end, like most of the popular deep-learning based crowd counting frameworks. Both part of this phase is trained simultaneously, using a weighted pixelwise MSE loss function given by:
\begin{displaymath}
	L\left( \Theta \right) = \frac{\sigma}{N}\sum_{i=1}^{N}{\norm{{M}_{{x}_{i}}^{\textrm{Pred}} \left( \Theta \right) - {M}_{{x}_{i}}^{\textrm{GT}} }_{2}^{2}}
\end{displaymath}
Here $\Theta$ is the set of parameters of the crowd model, N is total number of images, ${{M}_{{x}_{i}}^{\textrm{Pred}}}$ is the predicted density map of image, ${{M}_{{x}_{i}}^{\textrm{GT}}}$ is the ground truth map and $ \sigma $ denotes the weight. We found a weight of $3.5\times {10}^{4}$ quite satisfactory to speed up the training process.  While training the whole network, the output of the regression part is also optimized but at a discounted weight ($3.5\times {10}^{2}$) which is added to the total loss of the whole phases' network. We used Adam optimizer with a learning rate of $1\times {10}^{-6}$. On the whole training set, we used 1500 epochs with a batch size of 64 to optimize our model.

During the training process, we faced several saddle points in the weight space. To get rid of them, we checked for that by monitoring loss change for 15 epochs. Whenever we detect one, we increased the learning rate to $5\times {10}^{-4}$ to escape those points. This way we optimize our model to be adaptive in various environments and densities. %
\subsection{Cross-Branch Segregating Refinement}
This phase is the most difficult part to train in our whole network. Straightforward end-to-end training in this phase without any pre-training almost immediately falls into local minima resulting in early convergence. For the hyperparameter configuration of the layers of each branch, we started with the middle branch configuration of MCNN \cite{ref_single}, as we have already collected local information in earlier stages. We have tuned each branch until we get the best result on validation set.

To avoid the local minima, we first pre-train each of the 2 branches in this phase separately, upto the crossing stage. We use our own dataset to train this part of the network. During this pre-training, like in the second phase, we use the pixelwise MSE loss here too.
\begin{displaymath}
	L_{sit}\left( \Theta \right) = \frac{\sigma_{1}}{N}\sum_{i=1}^{N}{\norm{{M}_{{x}_{i}sit}^{\textrm{Pred}} \left( \Theta \right) - {M}_{{x}_{i}sit}^{\textrm{GT}} }_{2}^{2}}
\end{displaymath}

\begin{displaymath}
	L_{stand}\left( \Theta \right) = \frac{\sigma_{2}}{N}\sum_{i=1}^{N}{\norm{{M}_{{x}_{i}stand}^{\textrm{Pred}} \left( \Theta \right) - {M}_{{x}_{i}stand}^{\textrm{GT}} }_{2}^{2}}
\end{displaymath}
Here $ \sigma_1 $ and $ \sigma_2$ denotes the loss weights of sitting and standing branch. With a learning rate of $1\times {10}^{-5}$, these branches are fairly easy to train with 1000 epochs each. The best resulting model on the cross-validation dataset was selected. We now load those pre-trained weights and then add the cross connection on them. During the back-propagation of this joint training, loss of each branch is intermixed with the other branch in the crossing stage so that both branches can agree in terms of segregation. Here we allocate a slightly higher weight to $\sigma_1$, as equal weight allocation results in the whole architecture getting biased toward the standing network. This time we train the whole network with a learning rate of $1\times {10}^{-6}$ for 1000 epochs with 64 batch size. Thus with the help of pre-training, we can optimize this phase eliminating early convergence issues.

%% file: transfer/Experiments.tex
\begin{figure*}[t]
\centering
  \includegraphics[width=0.78\paperwidth, height=.21\paperheight]{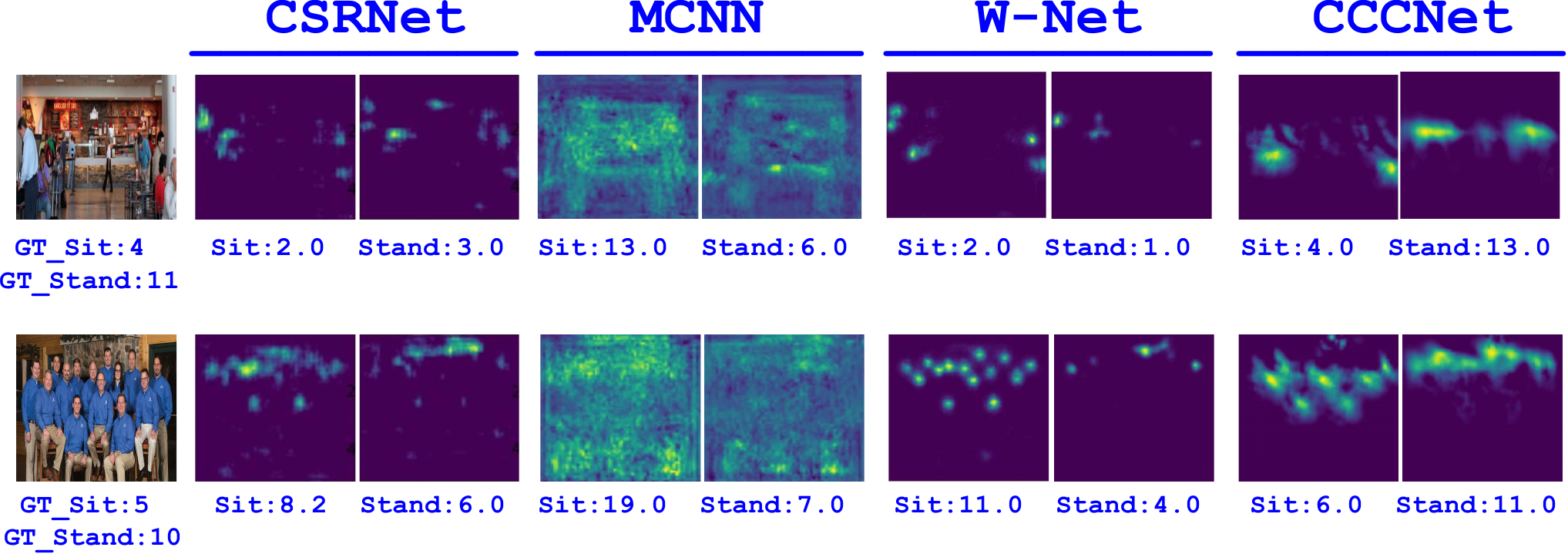}
  \caption{Sitting and standing result comparison on our dataset. Ground truth count is denoted under the original image, and predicted count of each model is denoted under respective density maps.}
  \label{fig:fig0}
\end{figure*}
We have conducted an extensive experimental study to show the efficacy of our approach.  As \textbf{there is no prior work} that directly answers the categorized counting problem, we cannot compare our solutions with a baseline from existing works. However, to compare the performance of our model with state-of-the-art methods in crowd counting, we formulate them as multi-task schemes for counting sitting and standing separately. As representatives of the state-of-the-art methods, we take MCNN \cite{ref_single} as representative of shallow multi-branch architectures due to its influential role in crowd counting, CSRNet \cite{li2018csrnet} as representative of deeper models due to its being the former benchmark leading performance, and state-of-the-art W-Net\cite{valloli2019w} architecture because of its high representational power and strong performance in raw crowd counting. %
\subsection{Dataset}
Crowd counting datasets are semantically rich containing a wide variety of information in a single image. This comes from the fact that each image has different crowd distributions at different parts of the image. In addition, crowd counting datasets are difficult to build up as each person in the image needs to be hand annotated separately. Consequently, they contain lesser no. of images than most other image datasets of other machine learning problem domains. Although there are a few benchmark crowd counting datasets used in previous crowd counting models (WorldExpo'10 Dataset \cite{ref_Idrees}, the ShanghaiTech Dataset \cite{ref_single}), none of these counting datasets can serve our purpose. Firstly, they do not contain enough number of sitting people as needed. No separate annotations for different categories are done either. Furthermore, the crowd density in most of the images in these datasets is so high that categorization, as sitting or standing, is of little use.

Therefore we build a new dataset%
of 553 crowd images randomly crawled from the Internet. These images contain from 1 to 206 persons per image with a total of 16521 persons taken from both indoor and outdoor environments with high illumination variation and filled with complex obstacles. We mainly focus on real-life environments with moderately high-density crowd where our system is most applicable. Few sample images of the dataset are shown in Figure \ref{fig_ds}. For each image, we separately annotate the head location of all the sitting and standing persons respectively. Also during training and validation, we perform horizontal flipping to further augment the data. 
\subsection{Results}
We show the categorization performance of the detected persons in the first phase in Table \ref{table_module1}. We use the traditional detection metrics of precision, recall, F1 score and combined accuracy to measure our performance. Standing is assumed to be the positive class and sitting posture to be the negative class. We achieve satisfactory categorization accuracy of detected persons of \textbf{86.50\%}. However, these categorization accuracies are based only on the persons detected by the pose estimator in the test images.
\begin{table}[h]
\begin{center}
\caption{\label{table_module1} Categorization performance of detected persons in the first phase}
\vspace{-2mm}
\begin{tabular}{|c|c|c|}
    \hline
    Category & Sitting & Standing\\
    \hline
    Precision(\%) & 87.47 & 85.20\\
    \hline
    Recall(\%) & 88.80 & 83.51\\
    \hline
    F1 Score(\%) & 88.13 & 84.35\\
    \hline
    Accuracy(\%) & \multicolumn{2}{|c|}{\textbf{86.499}}\\
    \hline
\end{tabular}
\end{center}
\end{table}
In Table \ref{detection_perf}, we show the mean absolute error (MAE) and root mean squared error (RMSE) of crowd count using faster RCNN detection framework  as used in \cite{ref_alphapose} in low (person count $<$ 25) and high person density images (person count $\geq$ 25) of our dataset. We observe that the detection errors rise significantly in higher density images. This poor human detection accuracy necessitates the second and third phases of our model to achieve an acceptable accuracy with higher density images.\\
\begin{table}[ht]
\begin{center}
\caption{\label{detection_perf}Detection performance}
\vspace{-2mm}
\begin{tabular}{|c|c|c|}
    \hline
    Metric & Low Density & High Density\\
    \hline
    Average Persons & 11.42 & 62.15\\
    \hline
    Detection MAE & 1.17 & 23.77\\
    \hline
    Detection RMSE & 2.04 & 31.05\\
    \hline
\end{tabular}
\end{center}
\end{table}
\begin{table*}[h]
\begin{center}
\caption{\label{overall}Overall performance comparison}
\vspace{-2mm}
\begin{tabular}{|c|c|c|c|c|}
    \hline
    Metric & \multicolumn{2}{|c|}{MAE} &  \multicolumn{2}{|c|}{MSE}\\
    \hline
    Method & Sitting & Standing & Sitting & Standing\\
    \hline
    MCNN & 13.89 & 9.029 & 17.54 & 16.399 \\
    \hline
    CSRNet & 7.33 & 6.51 & 13.28 & 11.61\\
    \hline
    W-Net & 6.38 & 6.95 & 11.63 & 15.01\\
    \hline
    SDBC (Only Phase 1) & 4.60 & 5.93 & 9.18 & 15.06\\
    \hline
    \textbf{CCCNet} & \textbf{4.15} & \textbf{4.80} & \textbf{7.96} & \textbf{8.59}\\
    \hline
    
\end{tabular}
\end{center}
\end{table*}
In Table \ref{overall}, we compare the performance of CCCNet with standalone detection based counting, SDBC (categorized count using the detection based density maps generated on the \textbf{first phase} of our pipeline in Section \ref{sec:DNN} ) and also with other state-of-the-art counting methods \cite{ref_single, li2018csrnet, valloli2019w} considering sitting and standing crowd counting as separate tasks. In all results, it is  evident that CCCNet significantly outperforms all  other models including current state-of-the-art counting models with a multi-task scheme adaptation. MCNN and CSRNet don't perform well in this case. While W-Net with its strong crowd counting scheme shows some reasonable performance, still it gives 53.7\% worse sitting MAE, 44.8\% worse standing MAE, 46.1\% worse sitting RMSE and 74.7\% worse standing RMSE than our proposed CCCNet. In fact, W-Net performs worse than the first phase of our pipeline alone in almost all the cases, as it only focuses on keeping the crowd count right, rather than being careful about categorization. This proves the effectiveness of CCCNet for categorized crowd counting over existing crowd counting approaches.

\subsection{Ablation Studies}
Our model primarily works by inferring attention maps from initial detection based density maps and total crowd map and later using those on them to get the final categorized maps. Leveraging this clever technique, CCCNet combines both local and global information necessary for categorized counting. This increases performance in every case over standalone detection based counting as evident in Table \ref{overall}. Some of the recent models \cite{li2018csrnet, gao2019scar, sam2017switching, liu2018leveraging} achieved exceptional performance in traditional crowd counting due to the usage of deep networks like VGG-16 \cite{simonyan2014very} or similar deeper architectures. While these helped to achieve high-quality density maps as well as more accurate counting, in our case where categorization is necessary they don't perform well. Except \cite{gao2019scar}, all the other networks suffer from low gradient which essentially leads to getting stuck in local optima as well as slow training. In Figure \ref{fig:fig0}, we see the performance of CSRNet as representative of the best performing deep architectures, generating poor density maps that do not resemble categorized crowd at all. They rather seem to be focusing on raw crowd count. Similar types of performances are found with other deeper networks \cite{sam2017switching, liu2018leveraging}. We observe that MCNN also ends up focusing on raw crowd count, and not doing a good job at that either. This behavior of leaning towards traditional crowd counting is most prominently seen with \cite{gao2019scar}, the present state-of-the-art and benchmark leading counting mechanism. This is a very powerful network and due to its reinforced U-Net like structure, it has lots of residual connections between final layers and initial layers along with its deep structure, giving it high representational power. These residual connections eliminate the problems with low gradient. With our dataset, we see this approach also focuses on producing crowd density map instead, generating high-quality \textit{traditional crowd density map}, that is the total crowd map without any categorization. Nevertheless, for sitting cases in Figure \ref{fig:fig0}, even though we have trained W-Net using only sitting annotations, we see it captures standing crowd too. This ultimately leads to crowd density maps that fail to capture categorization information. But here, our model successfully separates the sitting and standing crowd density map and gives commendable accuracy in giving the counts. This demonstrates the superiority of our model in categorized crowd counting over any other available counting techniques.

%% file: transfer/Conclusions.tex
In this paper, we have introduced a new form of crowd counting, namely categorized crowd counting, which counts the number of people sitting and standing in an image. To solve the categorized crowd counting problem, we propose a three phase deep learning architecture, \textit{CCCNet} that incorporates both detection based categorized density maps and global crowd density maps using attention mechanism to effectively count the number of people sitting and standing in an image. Extensive experiments on images of highly varying person densities and cross-scene environments show the effectiveness and superiority of CCCNet over other competitive techniques. On average, CCCNet only incurs a MAE of \textbf{4.15} and \textbf{4.80} and a RMSE of \textbf{7.96} and \textbf{8.59} for sitting and standing crowd count, respectively. In future, exploration of models generating higher resolution density maps may lead to even better categorized crowd counting performance.
\balance